\documentclass[11pt,a4paper]{article}
\usepackage[hyperref]{udst2018}
\usepackage{amsmath}
\usepackage{times}
\usepackage{latexsym}
\usepackage{subcaption}
\usepackage{graphicx}
\usepackage{url}

\DeclareMathOperator*{\argmax}{argmax}

\aclfinalcopy 
\def\aclpaperid{***} %  Enter the Paper ID here for final camera ready copy

\title{Towards Better UD Parsing: Deep Contextualized Word Embeddings, Ensemble, and Treebank Concatenation}

\author{Wanxiang Che, Yijia Liu, Yuxuan Wang, Bo Zheng, Ting Liu \\
	Research Center for Social Computing and Information Retrieval \\
	Harbin Institute of Technology, China \\
	{\tt \{car,yjliu,yxwang,bzheng,tliu\}@ir.hit.edu.cn}	}

\date{}
\begin{document}
\maketitle

\begin{abstract}
This paper describes our system (HIT-SCIR)
submitted to the  CoNLL 2018 shared
task on Multilingual Parsing from Raw Text to 
Universal Dependencies.
We base our submission on Stanford's winning system for the CoNLL 2017 shared task
and make two effective extensions: 
1) incorporating deep contextualized
word embeddings into both the part of speech
tagger and dependency parser;
2) ensembling parsers trained with different initialization.
We also explore different ways of concatenating treebanks
for further improvements.
Experimental results on the development data
show the effectiveness of our methods.
In the final evaluation,
our system was ranked first according to LAS (75.84\%)
and outperformed the other systems by a large margin.

\end{abstract}

\section{Introduction}

In this paper, we describe our system (HIT-SCIR) submitted to CoNLL 2018 shared
task on Multilingual Parsing from Raw Text to 
Universal Dependencies \cite{udst:overview}.
We base our system on Stanford's winning system \citep[\S\ref{sec:biaffine}]{dozat-qi-manning:2017:K17-3}
for the CoNLL 2017 shared task \cite{udst:overview2017}.

\citet{DBLP:journals/corr/DozatM16} and
its extension \cite{dozat-qi-manning:2017:K17-3} have
shown very competitive performance in both the shared task \cite{dozat-qi-manning:2017:K17-3}
and previous parsing works \cite{ma-hovy:2017:I17-1,shi-huang-lee:2017:EMNLP2017,N18-1088,DBLP:journals/corr/abs-1805-01087}.
A natural question that arises is how can we further improve
their part of speech (POS) tagger and dependency parser
via a simple yet effective technique.
In our system, we make two noteworthy extensions to their tagger and parser:
\begin{itemize}
	\item Incorporating the deep contextualized word embeddings \cite[ELMo: Embeddings from Language Models]{N18-1202} into the word representaton (\S\ref{sec:elmo});
	\item Ensembling parsers trained with different initialization (\S\ref{sec:ens}).
\end{itemize}
%Both these two extensions result in substantial improvements. 

For some languages in the shared task, multiple treebanks of different domains are provided.
Treebanks which are of the same language families are provided as well.
Letting these treebanks help each other has been shown an effective way to improve parsing performance
in both the cross-lingual-cross-domain parsing community and last year's shared tasks \cite{TACL892,guo-EtAl:2015:ACL-IJCNLP2,che-EtAl:2017:K17-3,shi-EtAl:2017:K17-3,bjorkelund-EtAl:2017:K17-3}.
In our system, we apply the simple concatenation to the treebanks that are potentially
helpful to each other
and explore different ways of concatenation to improve the parser's performance (\S\ref{sec:comb}).

In dealing with the small treebanks and treebanks from low-resource languages (\S\ref{sec:low}),
we adopt the word embedding transfer idea 
in the cross-lingual dependency parsing \cite{guo-EtAl:2015:ACL-IJCNLP2}
and use the bilingual word vectors transformation technique \cite{DBLP:journals/corr/SmithTHH17}\footnote{\url{https://github.com/Babylonpartners/fastText_multilingual}}
to map \textit{fasttext}\footnote{\url{https://github.com/facebookresearch/fastText}} word embeddings \cite{DBLP:journals/corr/BojanowskiGJM16}
of the source rich-resource language and target low-resource language
into the same space.
The transferred parser trained on the source language is used for the target low-resource language.

We conduct experiments on the development data to study
the effects of ELMo, parser ensemble, and treebank concatenation.
Experimental results show that these techniques substantially improve the parsing performance.
Using these techniques, our system achieved an averaged LAS of 75.84 on the official test set
and was ranked the first according to LAS \cite{udst:overview}.
This result significantly outperforms the others by a large margin.\footnote{\url{http://universaldependencies.org/conll18/results.html}}

We release our pre-trained ELMo for many languages at \url{https://github.com/HIT-SCIR/ELMoForManyLangs}.

\section{Deep Biaffine Parser}\label{sec:biaffine}

We based our system on the tagger and parser of \citet{dozat-qi-manning:2017:K17-3}.
The core idea of the tagger and parser
is using an LSTM network to produce the vector representation
for each word and then predict POS tags and dependency relations
using the representation.
For the tagger whose input is the word alone,
this representation is calculated as
\[
\mathbf{h}_i =  \text{BiLSTM}(\mathbf{h}_0, (\mathbf{v}_1^{(word)}, ..., \mathbf{v}_n^{(word)}))_i
\]
where $\mathbf{v}_i^{(word)}$ is the word embeddings.
After getting $\mathbf{h}_i$,
the scores of tags are calculated as
\begin{align*}
\mathbf{h}_i^{(pos)} & = \text{MLP}^{(pos)} (\mathbf{h}_i) \\
\mathbf{s}_i^{(pos)} & = W \cdot \mathbf{h}_i^{(pos)} + \mathbf{b}^{(pos)} \\
y_i^{(pos)} & = \argmax_{j} s_{i, j}^{(pos)} 
\end{align*}
where each element in $\mathbf{s}_i^{(pos)}$ represents the possibility
that $i$-th word is assigned with corresponding tag.

For the parser whose inputs are the word and POS tag,
such representation is calculated as
\begin{align*}
\mathbf{x}_i & =  \mathbf{v}_i^{(word)} \oplus \mathbf{v}_i^{(tag)} \\
\mathbf{h}_i & =  \text{BiLSTM}(\mathbf{h}_0, (\mathbf{x}_1, ..., \mathbf{x}_n))_i
\end{align*}
And a pair of representations are fed into a biaffine classifier
to predict the possibility that there is a dependency arc between these two words.
The scores over all head words are calculated as
\begin{align*}
\mathbf{s}_i^{(arc)} & = H^{(arc\text{-}head)} W^{(arc)} \mathbf{h}_i^{(arc\text{-}dep)} \\
& + H^{(arc\text{-}head)} \mathbf{b}^{(arc)} \\
y^{(arc)} & = \argmax_{j} s_{i, j}^{(arc)}
\end{align*}
where $\mathbf{h}_i^{(arc\text{-}dep)}$ is computed by feeding $\mathbf{h}_i$ into an MLP
and $H^{(arc\text{-}head)}$ is the stack of $\mathbf{h}_i^{(arc\text{-}head)}$
which is calculated in the same way as $\mathbf{h}_i^{(arc\text{-}dep)}$
but using another MLP.
After getting the head $y^{(arc)}$ word,
its relation with $i$-th word is decided by calculating 
\begin{align*}
\mathbf{s}_i^{(rel)} & = \mathbf{h}^{T(rel-head)}_{y^{‘(arc)}} \mathbf{U}^{(rel)} \mathbf{h}_i^{(rel-dep)} \\
& + W^{(rel)} (\mathbf{h}_i^{(rel-dep)} \oplus \mathbf{h}^{T(rel-head)}_{y^{‘(arc)}}) \\
& + \mathbf{b}^{(rel)}, \\
y^{(rel)} & = \argmax_{j} s_{i, j}^{(rel)}
\end{align*}
where $\mathbf{h}^{(rel-head)}$ and $\mathbf{h}^{(rel-dep)}$
are calculated in the same way as $\mathbf{h}_i^{(arc\text{-}dep)}$ and $\mathbf{h}_i^{(arc\text{-}head)}$.

This decoding process can lead to cycles in the result.
\citet{dozat-qi-manning:2017:K17-3} employed an iterative fixing methods on the cycles.
We encourage the reader of this paper to refer to their paper for more details on training and decoding.

For both the biaffine tagger and parser, 
the word embedding $\mathbf{v}_i^{(word)}$ is obtained by summing 
a fine-tuned token embedding $\mathbf{w}_i$, a fixed word2vec embedding $\mathbf{p}_i$, and an LSTM-encoded
character representation $\mathbf{\hat{v}}_i$ as
\[
\mathbf{v}_i^{(word)} = \mathbf{w}_i + \mathbf{p}_i + \mathbf{\hat{v}}_i.
\]

\section{Deep Contextualized Word Embeddings}\label{sec:elmo}

Deep contextualized word embeddings \citep[ELMo]{N18-1202}
has shown to
be very effective on a range of syntactic and semantic tasks
and it's straightforward to obtain ELMo by
using an LSTM network to encode words in a sentence
and training the LSTM network with language modeling objective
on large-scale raw text.
More specifically, the $\mathbf{ELMo}_i$ is computed
by first computing the hidden representation $\mathbf{h}_i^{(LM)}$ as
\[
\mathbf{h}_i^{(LM)} =  \text{BiLSTM}^{(LM)}(\mathbf{h}_0^{(LM)}, (\mathbf{\tilde{v}}_1, ..., \mathbf{\tilde{v}}_n))_i
\]
where $\mathbf{\tilde{v}}_i$ is the output of a CNN over characters,
then attentively summing and scaling different layers of  $\mathbf{h}_{i, j}^{(LM)}$
with $s_j$ and $\gamma$
as
\[
\mathbf{ELMo}_i = \gamma \sum_{j=0}^L s_j \mathbf{h}_{i, j}^{(LM)},
\]
where $L$ is the number of layers and $\mathbf{h}_{i, 0}^{(LM)}$ is identical to $\mathbf{\tilde{v}}_i$.
In our system, we follow \citet{N18-1202} and use a two-layer bidirectional LSTM as our $\text{BiLSTM}^{(LM)}$.

In this paper, we study the usage of ELMo for improving both
the tagger and parser and make several simplifications.
Different from \citet{N18-1202}, we treat the output of ELMo as a fixed representation
and do not tune its parameters during tagger and parser training.
Thus, we cancel the layer-wise attention scores $s_j$ and the scaling factor $\gamma$, 
which means
\[
\mathbf{ELMo}_i = \sum_{j=0}^{2} \mathbf{h}_{i, j}^{(LM)}.
\]
In our preliminary experiments, using $ \mathbf{h}_{i, 0}^{(LM)}$ for $\mathbf{ELMo}_i$
yields better performance on some treebanks.
In our final submission, we decide using either $\sum_{j=0}^{2} \mathbf{h}_{i, j}^{(LM)}$
or $ \mathbf{h}_{i, 0}^{(LM)}$ based on their development.

After getting $\mathbf{ELMo}_i$, we project it
to the same dimension as $\mathbf{v}_i^{(word)}$ and
use it as an additional word embedding.
The calculation of $\mathbf{v}_i^{(word)}$ becomes
\[
\mathbf{v}_i^{(word)} = \mathbf{w}_i + \mathbf{p}_i + \mathbf{\hat{v}}_i + W^{(ELMo)} \cdot \mathbf{ELMo}_i
\]
for both the tagger and parser.
We need to note that training the tagger and parser includes $W^{(ELMo)}$.
To avoid overfitting, we impose a dropout function on projected vector
$W^{(ELMo)} \cdot \mathbf{ELMo}_i$
during training.

\section{Parser Ensemble}\label{sec:ens}

%We seek to use ensemble to improve the performance.
According to \citet{reimers-gurevych:2017:EMNLP2017}, neural network training can
be sensitive to initialization and
\citet{DBLP:journals/corr/abs-1805-11224}  shows that ensemble
neural network trained with different initialization
leads to performance improvements.
We follow their works and train three parsers with different initialization,
then ensemble these parsers by averaging their softmaxed output scores as 
\[\mathbf{s}_i^{(rel)} = \frac{1}{3} \sum_{m=1}^{3} \text{softmax}(\mathbf{s}_i^{(m, rel)}).\]

\section{Treebank Concatenation}\label{sec:comb}

For 15 out of the 58 languages in the shared task,
multiple treebanks from different domains are provided.
There are also treebanks that come from the same language family.
Taking the advantages of the relation between treebanks has been shown 
a promising direction in both the research community \cite{TACL892,guo-EtAl:2015:ACL-IJCNLP2,C16-1002}
and in the CoNLL 2017 shared task \cite{che-EtAl:2017:K17-3,bjorkelund-EtAl:2017:K17-3,shi-EtAl:2017:K17-3}.
In our system, we adopt the treebank concatenation technique as \citet{TACL892}
with one exception: only a group of treebanks from the same language (\textit{cross-domain concatenation})
or a pair of treebanks that are typologically or geographically correlated (\textit{cross-lingual concatenation})
is concatenated.

In our system, we tried cross-domain concatenation on
\textit{nl}, \textit{sv}, \textit{ko}, \textit{it}, \textit{en}, \textit{fr},
\textit{gl}, \textit{la}, \textit{ru}, and \textit{sl}.\footnote{We opt out \textit{cs}, \textit{fi}, and \textit{pl} because all the treebanks of these languages are relatively large -- they have more than 10K training sentences.}
We also tried cross-lingual concatenation on \textit{ug-tr}, \textit{uk-ru}, \textit{ga-en},
and \textit{sme-fi} following \citet{che-EtAl:2017:K17-3}.
However, due to the variance in vocabulary, grammatical genre, and even annotation, 
treebank concatenation does not guarantee to improve the model's performance.
We decide the usage of concatenation by examining their development set performance.
For some small treebanks which do not have development set, whether using treebank concatenation
is decided through 5-fold cross validation.\footnote{We use \textit{udpipe}
	for this part of experiments
	because we consider the effect of treebank concatenation as
	being irrelevant to the parser architecture
	and \textit{udpipe} has the speed advantage in both training and testing.}
We show the experimental results of treebank concatenation
in Section \ref{sec:treebank-concat}.

%
%\subsection{Cross-Domain Concatenation}
%
%We test the performance of treebank concatenation.

%Based on the results in Table \ref{tbl:confuse-mat} and Table \ref{tbl:confuse-mat2},
%we apply cross-domain concatenation to the treebank where concatenation
%improves either the development or cross validation performance.
%
%\subsection{Cross-Lingual Concatenation}
%

\section{Low Resources Languages}\label{sec:low}

In the shared task, 5 languages are presented with training set of less than 50 sentences.
4 languages do not even have any training data.
It's difficult to train reasonable parser on these low-resource languages.
We deal with these treebanks by adopting the word embedding transfer idea 
of \citet{guo-EtAl:2015:ACL-IJCNLP2}.
We transfer the word embeddings of the rich-resource language
to the space of low-resource language using the bilingual word vectors transformation technique
\cite{DBLP:journals/corr/SmithTHH17}
and trained a parser using the source treebank
with only pretrained word embeddings on the transformed space
as $\mathbf{v}_i^{(word)} = \mathbf{p}_i$.
The transformation matrix is automatically learned on the \textit{fasttext} word embeddings
using the same tokens shared by two languages (like punctuation).

\begin{table}[t]
	\centering
	\small
	\begin{tabular}{r|cccccccc}
		\textit{target} & br & fo & th &hy & kk & bxr & kmr & hsb \\
		\hline
		\textit{source} & ga & no & zh & et & tr &  hi & fa & pl \\
	\end{tabular}
	\caption{Cross-lingual transfer settings for low-resource target languages.}\label{tbl:low-res-trans}
\end{table}
Table \ref{tbl:low-res-trans} shows our source languages for the target low-resource languages.
For a treebank with a few training data, its  source language is decided by
testing the source parser's performance on the training data.\footnote{We use \textit{udpipe} for this test.
	When training the parser, the small set of target training data is also used.}
For a treebank without any training data, we choose the source language according to their language family.\footnote{Thai
	does not have a treebank in the same family. We choose Chinese as source language because of geographical closeness
	and both these two languages are SVO in typology.}

\textit{Naija} presents an exception for our method since it does not have \textit{fasttext}
word embeddings and embedding transformation is infeasible.
Since it's a dialect of English, we use the full pipeline of \textit{en\_ewt} for \textit{pcm\_nsc} instead.
\begin{figure*}[t]
	\begin{subfigure}{\textwidth}
	\includegraphics[width=\textwidth]{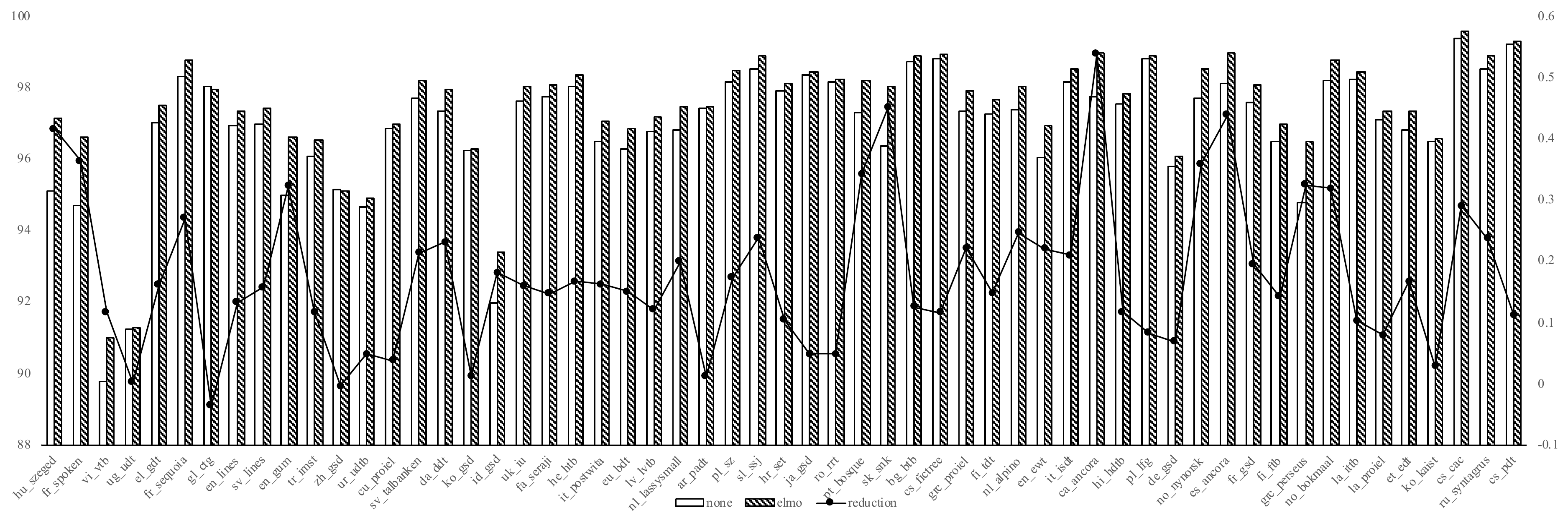}
	\caption{The effects of ELMo on POS tagging}\label{fig:elmo-effect:pos}
	\end{subfigure}
	\begin{subfigure}{\textwidth}
	\includegraphics[width=\textwidth]{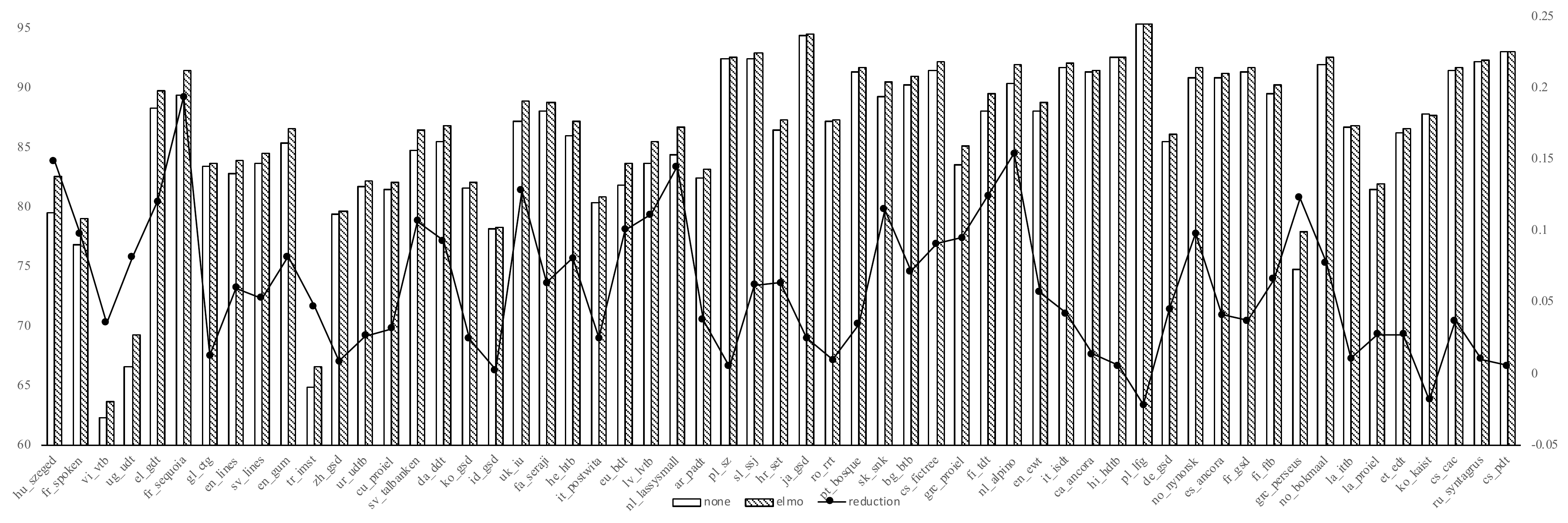}
	\caption{The effects of ELMo on dependency parsing}\label{fig:elmo-effect:par}
	\end{subfigure}
	\caption{The effects of ELMo.
		Treebanks are sorted from the smallest to the largest.}\label{fig:elmo-effect}
\end{figure*}

\section{Preprocessing}
Besides improving the tagger and parser,
we also consider the preprocessing as an important factor to
the final performance and improve it
by using the state-of-the-art system for sentence segmentation,
or developing our own word segmentor for languages
whose tokenizations are non-trival.

\subsection{Sentence Segmentation}
For some treebanks, sentence segmentation can be problematic
since there is no explicitly sentence delimiters.
\citet{delhoneux-EtAl:2017:K17-3} and \citet{DBLP:journals/corr/abs-1709-03756}
presented a joint tokenization and
sentence segmentation model (denoted as \textit{Uppsala segmentor})\footnote{\url{https://github.com/yanshao9798/segmenter/}}
that outperformed the baseline model in last year's shared task \cite{udst:overview2017}.
We select a set of treebanks whose \textit{udpipe} sentence segmentation F-scores
are lower than 95 on the development set and use Uppsala segmentor instead.\footnote{We
	use Uppsala segmentor for \textit{it\_postwita}, \textit{got\_proiel}, \textit{la\_poroiel}, \textit{cu\_proiel},
	\textit{grc\_proiel}, \textit{sl\_ssj}, \textit{nl\_lassysmall}, \textit{fi\_tdt}, \textit{pt\_bosque}, \textit{da\_ddt}, \textit{id\_gsd},
	\textit{el\_gdt}, and \textit{et\_edt}.}
Using the Uppsala segmentor leads to a development improvement of 
7.67 F-score in these treebanks over \textit{udpipe} baseline
and it was ranked the first according to sentence segmentation
in the final evaluation.

\subsection{Tokenization for Chinese, Japanese, and Vietnamese}

Tokenization is non-trivial for languages 
which do not have explicit word boundary markers, like Chinese, Japanese, and Vietnamese.
We develop our own tokenizer (denoted as \textit{SCIR tokenizer}) for these three languages.
Following \citet{che-EtAl:2017:K17-3} and \citet{10.1007/978-3-319-69005-6_6}, we model the tokenization as labeling the
word boundary tag\footnote{We use the BIES scheme.} on characters and 
use features derived from large-scale unlabeled data to further improve the performance.\footnote{For Vietnamese where whitespaces occur both inter- and intra-words, we treat the whitespace-separated token as a character.}
In addition to the pointwise mutual information (PMI), we also incorporate
the character ELMo into our tokenizer.
Embeddings of these features are concatenated along with a bigram character embeddings
as input.
These techniques lead to the best tokenization performance on all the related treebanks
and the average improvement over \textit{udpipe} baseline is 7.5 in tokenization F-score.\footnote{on \textit{ja\_gsd}, \textit{ja\_modern}, \textit{vi\_vtb}, and \textit{zh\_gsd}.}

\subsection{Preprocessing for Thai}

Thai language presents a unique challenge in the preprocessing.
Our survey on the Thai Wikipedia indicates that there is no explicit sentence delimiter 
and obtaining Thai words requires tokenization.
To remedy this, we use the whitespace as sentence delimiter and
use the lexicon-based word segmentation -- forward maximum matching algorithm
for Thai tokenization.
Our lexicon is derived from the \textit{fasttext} word embeddings by preserving the top 10\% frequent words.

\subsection{Lemmatization and Morphology Tagging}
We did not make an effort on lemmatization and morphology tagging,
but only use the baseline model.
This lags our performance in the MLAS and BLEX evaluation,
in which we were ranked 6th and 2nd correspondingly.
However, since our method, especially incorporating ELMo, is not limited to particular task,
we expect it to improve both the lemmatization and morphology tagging and
achieve better MLAS and BLEX scores.

\section{Implementation Details}
\begin{figure*}[t]
	\includegraphics[width=\textwidth]{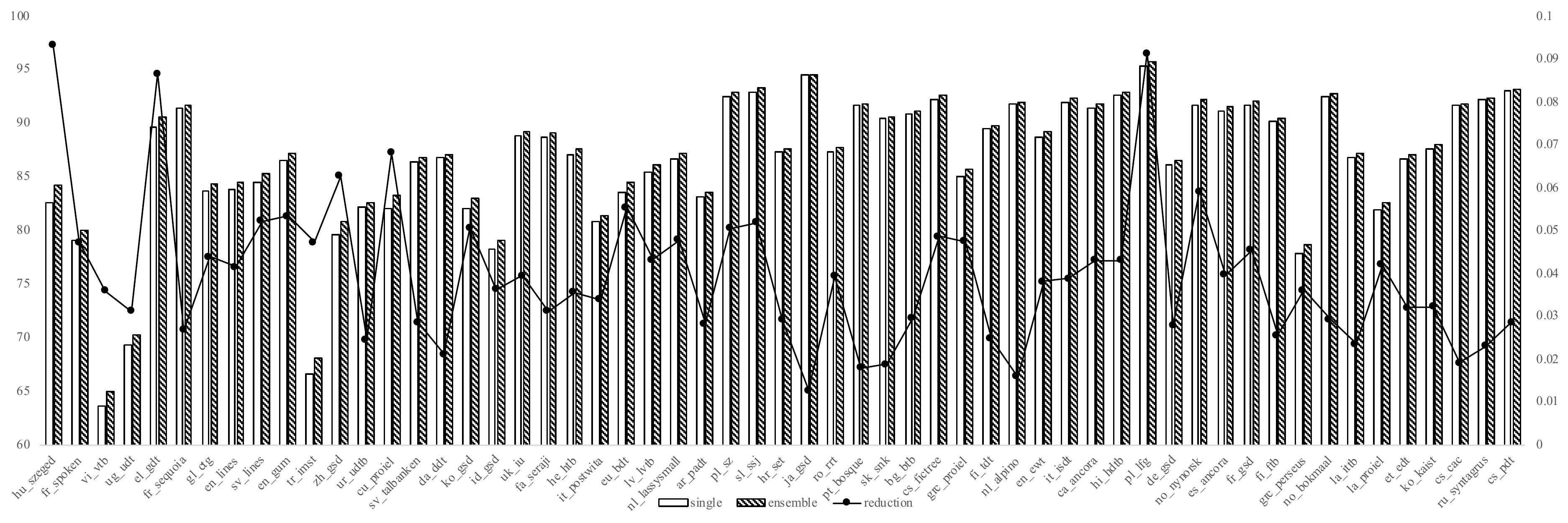}
	\caption{The effects of ensemble on dependency parsing.
		Treebanks are sorted according to the number of training sentences from left to right.}\label{fig:elmo-effect-ens}
\end{figure*}
\begin{table*}[t]
	\centering
	\small
	\setlength{\tabcolsep}{5pt}
	\begin{tabular}{rcc || rcc || rcc || rcc}
		%	\hline
		\textit{nl} & apino & lassysmall & \textit{sv} & lines & talbanken & \textit{ko} & gsd & kaist & \textit{it} & isdt & postwita \\
		\# train & 12.2 & 5.8 & \# train & 2.7 & 4.3 & \# train & 4.4 & 23.0 & \# train & 13.1 & 5.4 \\
		\hline
%		apino & 91.87 & & lines & 84.64 & & gsd & 82.05 & & isdt & \textbf{92.01} & \\
%		lassysmall & & 86.82 & talbanken & & 86.39 & kaist & & \textbf{87.83} & postwita & & 80.79 \\
		single & 91.87 & 86.82 & single & 84.64 & 86.39 & single & 82.05 & \textbf{87.83} & single & \textbf{92.01} & 80.79 \\
		%\hline
		concat. & \textbf{92.08} & \textbf{89.34} & concat. & \textbf{85.76} & \textbf{86.77} & concat. & \textbf{83.73} & 87.61 & concat.& 91.80 & \textbf{82.54} \\
		%	\hline
		\vspace*{0.5em}
	\end{tabular}
	\begin{tabular}{rccc || rccc}
		\textit{en} & ewt & gum & lines & \textit{fr} & gsd & sequoia & spoken \\
		\# train & 12.5 & 2.9 & 2.7 & \# train & 14.6 & 2.2 & 1.2\\
		\hline
%		ewt & \textbf{88.75} & & & gsd & \textbf{91.64} & & \\
%		gum & & \textbf{86.52} & & sequoia & & \textbf{91.44} & \\
%		lines & & & 83.86 & spoken & & & 79.06 \\
		single & \textbf{88.75} & \textbf{86.52} & 83.86 & single &\textbf{91.64} & \textbf{91.44} & 79.06 \\
		%\hline
		concat. & 88.74 & 85.65 & \textbf{85.30} & concat. & 91.44 & 90.51 & \textbf{81.99} \\
	\end{tabular}
	\caption{The developement performance with cross-domain concatenation for languages which has multiple treebanks.
		\textit{single} means training the parser on it own treebank without concatenation.
		\textit{\# train} shows the number of training sentences in the treebank measured in thousand.}\label{tbl:confuse-mat}
\end{table*}

\paragraph{Pretrained Word Embeddings.}
We use the 100-dimensional pretrained word embeddings released by the shared task
for the large languages.
For the small treebanks and treebanks for low-resource languages where cross-lingual transfer is required,
we use the 300-dimensional \textit{fasttext} word embeddings.
Old French treebank (\textit{fro\_srcmf}) presents the only exceptions and we use the French embeddings instead.
For all the embeddings, we only use 10\% of the most frequent words.

\paragraph{ELMo.}
We use the same hyperparameter settings as \citet{N18-1202} for $\text{BiLSTM}^{(LM)}$
and the character CNN.
We train their parameters
as training a bidirectional language model
on a set of 20-million-words data randomly
sampled from the raw text released by the shared task for each language.
Similar to \citet{N18-1202}, we use the \textit{sample softmax} technique
to make training on large vocabulary feasible \cite{jean-EtAl:2015:ACL-IJCNLP}.
However, we use a window of 8192 words surrounding the target word
as negative samples and it shows better performance in our preliminary experiments.
The training of ELMo on one language takes roughly 3 days on an NVIDIA P100 GPU.

\paragraph{Biaffine Parser.}
We use the same hyperparameter settings
as \citet{dozat-qi-manning:2017:K17-3}.
When trained with ELMo, we use a dropout of 33\% on the projected vectors.

\paragraph{SCIR Tokenizer.}
We use a
50-dimensional character bigram embeddings.
For the character ELMo whose input is a character,
the language model predict next character  in the same way as the word ELMo.
The final model is an ensemble of five single segmentors.

\paragraph{Uppsala Segmentor.}
We use the default settings for the Uppsala segmentor and the
final model is an ensemble of three single segmentors.

\begin{table*}[t]
	\centering
	\small
	\begin{tabular}{rc || rc || rc || rc || rc}
		\textit{gl} & treegal & \textit{la} & perseus & \textit{no} & nynorsklia & \textit{ru} & taiga & \textit{sl} & sst \\
		\# train & 0.6 & \# train & 1.3 & \# train & 0.3 & \# train & 0.9 & \# train & 2.1 \\
		\hline
		treegal & \textbf{66.71} & perseus & 44.05 & nynorsklia & 51.05 & taiga & 54.70 & sst & 55.15 \\
		+ctg & 56.73 & +proiel & \textbf{50.78} & +nynorsk & \textbf{58.49} & +syntagrus & \textbf{60.75} & +ssj & \textbf{59.52} \\
	\end{tabular}
	\caption{The 5-fold cross validation results for the cross-domain concatenation of treebank which does not have development set.}\label{tbl:confuse-mat2}
\end{table*}
\begin{table*}[t]
	\centering
	\small
	\begin{tabular}{rc || rc || rc || rc}
		& ug\_udt  &  & uk\_iu & & ga\_idt & & sme\_giella \\
		\hline
		ug\_udt & \textbf{69.27} & uk\_iu & 88.84 & ga\_idt & \textbf{62.84} & sme\_giella & \textbf{66.33}\\
		+tr\_imst & 19.27 & +ru\_syntagus & \textbf{90.74} & +en\_ewt &51.00 & +fi\_ftb & 59.86\\
	\end{tabular}
	\caption{Cross-lingual concatenation results. 
		The results for \textit{ug\_udt} and \textit{uk\_iu} are obtained on the development set.
		The results for \textit{ga\_idt} and \textit{sme\_giella} are obtained with \textit{udpipe} by 5-fold cross validation.}\label{tbl:cross-ling-concat}
\end{table*}

\section{Results}

\subsection{Effects of ELMo}

We study the effect of ELMo on the large treebanks and report
the results of a single tagger and parser with and without ELMo.
Figure \ref{fig:elmo-effect:pos} shows the tagging results on the development set
and Figure \ref{fig:elmo-effect:par}
shows the parsing results.
Using ELMo in the tagger leads to a macro-averaged improvement of 0.56\% in UPOS
and the macro-averaged error reduction is 17.83\%.
Using ELMo in the parser leads to a macro-averaged improvement 
of 0.84\% in LAS and
the macro-averaged error reduction is 7.88\%.

ELMo improves the tagging performance almost on every treebank,
except for \textit{zh\_gsd} and \textit{gl\_ctg}.
Similar trends are witnessed in the parsing experiments with \textit{ko\_kaist}
and \textit{pl\_lfg} being the only treebanks where ELMo slightly worsens the performance.

We also study the relative improvements in dependence on the size of the treebank.
The line in Figure \ref{fig:elmo-effect:pos} and Figure \ref{fig:elmo-effect:par}
shows the error reduction from using ELMo on each treebank.
However, no clear relation is revealed between the treebank size and the gains using ELMo.

\subsection{Effects of Ensemble}

We also test the effect of ensemble and show
the results in Figure \ref{fig:elmo-effect-ens}.
Parser ensemble leads to an averaged improvement of 0.55\% in LAS
and the averaged error reduction is 4.0\%.
These results indicate that ensemble is an effective way to
improve the parsing performance.
The relationship between gains using ensemble and treebank size
is also studied in this figure and the trend is that small treebank benefit more
from the ensemble.
We address this to the fact that the ensemble improves the model's generalization
ability in which the parser trained on small treebank is weak due to overfitting.

\subsection{Effects of Treebank Concatenation}\label{sec:treebank-concat}
As mentioned in Section \ref{sec:comb},
we study the effects of both the \textit{cross-domain concatenation} and \textit{cross-lingual concatenation}.
\paragraph{Cross-Domain Concatenation.}

For the treebanks which have development set, the development performances
are shown in Table \ref{tbl:confuse-mat}.
Numbers of sentences in the training set are also shown in this table.
The general trend is that for the treebank with small training set,
cross-domain concatenation achieves better performance.
While for those with large training set, concatenation does not improve
the performance or even worsen the results.

For the small treebanks which do not have development set,
the 5 fold cross validation results are shown in Table \ref{tbl:confuse-mat2}
in which concatenation improves most of the treebanks except for \textit{gl\_treegal}.

\paragraph{Cross-Lingual Concatenation.}

The experimental results of cross-lingual concatenation are shown in Table \ref{tbl:cross-ling-concat}.
Unfortunately, concatenating treebanks from different languages only
achieves improved performance on \textit{uk\_iu}.
This results also indicate that in cross lingual parsing,
sophisticated methods like word embeddings transfer \cite{guo-EtAl:2015:ACL-IJCNLP2,guo2016representation} and treebank transfer \cite{C16-1002}
are still necessary.

\subsection{Effects of Better Preprocessing}
\begin{table}[t]
  \centering
  \small
  \begin{tabular}{rccc}
     & $\Delta$-sent. & \textit{udpipe} & \textit{uppsala} \\
    \hline
    fi\_tdt & +0.69 & 88.13 & \textbf{88.67} \\
    et\_edt & +1.22 & 86.33 & \textbf{86.36} \\
    nl\_lassysmall & +1.39 & 88.08 & \textbf{88.60} \\
    da\_ddt & +1.56 & 86.21 & \textbf{86.51} \\
    el\_gdt & +1.57 & \textbf{90.08} & 89.96  \\
    cu\_proiel & +1.72 & 72.79 & \textbf{74.04} \\
    pt\_bosque & +1.83 & \textbf{90.73} & 90.20 \\
    id\_gsd & +2.46 & 74.14 & \textbf{78.83} \\
    la\_proiel & +4.82 & 73.21 & \textbf{74.22} \\
    got\_proiel & +5.36 & 67.55 & \textbf{68.40} \\
    grc\_proiel & +5.86 & 79.67 & \textbf{80.72} \\
    sl\_ssj & +18.81& 88.43 & \textbf{92.27} \\
    it\_postwita & +30.40 &74.91 & \textbf{79.26} \\
    \hline
    \hline
    & $\Delta$-word & \textit{udpipe} & \textit{scir} \\
    ja\_gsd & +4.07 & 80.53 & \textbf{85.23} \\
    zh\_gsd & +7.16 & 66.16 & \textbf{75.78} \\
    vi\_vtb & +9.02 & 48.58 & \textbf{57.53} \\
  \end{tabular}
\caption{The effects of improved preprocessing on the parsing performance.
	The first block shows the effects of sentence segmentation improvement.
	$\Delta$-sent. means the sentence segmentation F-score  difference between Uppsala segmentor and \textit{udpipe}.
	The second block shows the effects of word segmentation improvement.
	$\Delta$-word means the word segmentation in F-score difference  between SCIR tokenizer and \textit{udpipe}.}\label{tbl:preprocess}
\end{table}

\begin{table*}[t]
	\scriptsize
	\centering
	\setlength{\tabcolsep}{4pt}
	\begin{tabular}{rlllccc}
		\textit{ltcode} & sent+tokenize & tagger & parser & LAS &  w/o ens. & ref. LAS  \\
		\hline
		af\_afribooms & udpipe: self & biaffine (none): self & biaffine (none)*3: self & 85.47 (1) & 84.41 (5) & 85.45 \\
		ar\_padt & udpipe: self & biaffine ($h_{0,1,2}$): self & biaffine ($h_{0,1,2}$)*3: self & 73.63 (2) & 73.34 (3) & 77.06 \\
		bg\_btb & udpipe: self & biaffine ($h_{0}$): self & biaffine ($h_{0}$)*3: self & 91.22 (1) & 90.89 (1) & 90.41 \\
		br\_keb & udpipe: self & biaffine\_trans: self+ga\_idt & biaffine\_trans*3: self+ga\_idt & 8.54 (21)& 7.82 (21)& 38.64 \\
		bxr\_bdt & udpipe: self & biaffine\_trans: self+hi\_hdtb & biaffine\_trans*3: self+hi\_hdtb & 15.44 (6) & 15.69 (6)& 19.53  \\
		ca\_ancora & udpipe: self & biaffine ($h_{0}$): self & biaffine ($h_{0,1,2}$)*3: self & 91.61 (1) & 91.29 (1) & 90.82  \\
		cs\_cac & udpipe: self & biaffine ($h_{0}$): self & biaffine ($h_{0}$)*3: self & 91.61 (1) & 91.33 (1)& 91.00 \\
		cs\_fictree & udpipe: self & biaffine ($h_{0,1,2}$): self & biaffine ($h_{0,1,2}$)*3: self & 92.02 (1) & 91.39 (3) & 91.83 \\
		cs\_pdt & udpipe: self & biaffine ($h_{0}$): self & biaffine ($h_{0,1,2}$)*3: self & 91.68 (1) & 91.45 (1) & 90.57 \\
		cs\_pud & udpipe: cs\_pdt & biaffine ($h_{0}$): cs\_pdt & biaffine ($h_{0,1,2}$)*3: cs\_pdt & 86.13 (1) & 85.89 (1) & 85.35 \\
		cu\_proiel & uppsala: self & biaffine ($h_{0,1,2}$): self & biaffine ($h_{0,1,2}$)*3: self & 74.29 (3) & 73.29 (4) & 75.73  \\
		da\_ddt & uppsala: self & biaffine ($h_{0}$): self & biaffine ($h_{0,1,2}$)*3: self & 86.28 (1) & 85.54 (1) & 84.88 \\
		de\_gsd & udpipe: self & biaffine ($h_{0}$): self & biaffine ($h_{0,1,2}$)*3: self & 80.36 (1) & 79.81 (1) & 79.03 \\
		el\_gdt & uppsala: self & biaffine ($h_{0}$): self & biaffine ($h_{0,1,2}$)*3: self & 89.65 (1) & 88.88 (3) & 89.59 \\
		en\_ewt & udpipe: self & biaffine ($h_{0}$): self & biaffine ($h_{0,1,2}$)*3: self & 84.57 (1) & 83.88 (2) & 84.02  \\
		en\_gum & udpipe: self & biaffine ($h_{0}$): self & biaffine ($h_{0,1,2}$)*3: self & 84.42 (2) & 83.57 (2) & 85.05  \\
		en\_lines & udpipe: self & biaffine ($h_{0,1,2}$): self & biaffine ($h_{0}$)*3: self+en\_ewt+en\_gum & 81.97 (1) & 81.67 (1) & 81.44 \\
		en\_pud & udpipe: en\_ewt & biaffine ($h_{0}$): en\_ewt & biaffine ($h_{0,1,2}$)*3: en\_ewt & 87.73 (2) & 87.26 (2) & 87.89  \\
		es\_ancora & udpipe: self & biaffine ($h_{0}$): self & biaffine ($h_{0,1,2}$)*3: self & 90.93 (1) & 90.62 (1) & 90.47 \\
		et\_edt & uppsala: self & biaffine ($h_{0}$): self & biaffine ($h_{0,1,2}$)*3: self & 85.35 (1) & 84.74 (1) & 84.15 \\
		eu\_bdt & udpipe: self & biaffine ($h_{0,1,2}$): self & biaffine ($h_{0}$)*3: self & 84.22 (1) & 83.42 (1) & 83.13  \\
		fa\_seraji & udpipe: self & biaffine ($h_{0,1,2}$): self & biaffine ($h_{0,1,2}$)*3: self & 88.11 (1) & 87.60 (1) & 86.18  \\
		fi\_ftb & udpipe: self & biaffine ($h_{0,1,2}$): self & biaffine ($h_{0}$)*3: self & 88.53 (1) & 88.00 (1) & 87.86 \\
		fi\_pud & udpipe: fi\_tdt & biaffine ($h_{0}$): fi\_tdt & biaffine ($h_{0}$)*3: fi\_tdt & 90.23 (1) & 89.58 (1) & 89.37\\
		fi\_tdt & uppsala: self & biaffine ($h_{0}$): self & biaffine ($h_{0}$)*3: self & 88.73 (1) & 88.68 (1) & 87.64  \\
		fo\_oft & udpipe: no\_bokmaal & biaffine\_trans: no\_bokmaal & biaffine\_trans*3: no\_bokmaal & 44.05 (4)& 44.17 (4) & 49.43 \\
		fr\_gsd & udpipe: self & biaffine ($h_{0}$): self & biaffine ($h_{0,1,2}$)*3: self & 86.89 (1) & 86.81 (1) & 86.46 \\
		fr\_sequoia & udpipe: self & biaffine ($h_{0,1,2}$): self & biaffine ($h_{0,1,2}$)*3: self & 89.65 (2) & 89.12 (2) & 89.89\\
		fr\_spoken & udpipe: self & biaffine ($h_{0}$): self & biaffine ($h_{0,1,2}$)*3: self+fr\_gsd+fr\_sequoia & 75.78 (1) & 75.09 (1) & 74.31 \\
		fro\_srcmf & udpipe: self & biaffine (none): self & biaffine (none)*3: self & 87.07 (2) & 86.53 (3) & 87.12 \\
		ga\_idt & udpipe: self & biaffine (none): self & biaffine (none)*3: self & 68.57 (5) & 66.80 (7) & 70.88\\
		gl\_ctg & udpipe: self & biaffine (none): self & biaffine (none)*3: self & 82.35 (2) & 81.80 (3) & 82.76 \\
		gl\_treegal & udpipe: self & biaffine (none): self & biaffine (none)*3: self & 72.88 (4) & 71.27 (8) & 74.25 \\
		got\_proiel & uppsala: self & biaffine (none): self & biaffine (none)*3: self & 69.26 (3) & 67.61 (5) & 69.55 \\
		grc\_perseus & udpipe: self & biaffine ($h_{0,1,2}$): self & biaffine ($h_{0,1,2}$)*3: self & 79.39 (1) & 78.53 (1) & 74.29 \\
		grc\_proiel & uppsala: self & biaffine ($h_{0}$): self & biaffine ($h_{0,1,2}$)*3: self & 79.25 (1) & 78.35 (1) & 76.76 \\
		he\_htb & udpipe: self & biaffine ($h_{0}$): self & biaffine ($h_{0}$)*3: self & 67.05 (3) & 66.67 (3) & 76.09 \\
		hi\_hdtb & udpipe: self & biaffine ($h_{0,1,2}$): self & biaffine ($h_{0}$)*3: self & 92.41 (1) & 92.13 (1) & 91.75 \\
		hr\_set & udpipe: self & biaffine ($h_{0}$): self & biaffine ($h_{0,1,2}$)*3: self & 87.36 (1) & 86.82 (1) & 86.76 \\
		hsb\_ufal & udpipe: self & biaffine\_trans: self+pl\_lfg & biaffine\_trans*3: self+pl\_lfg & 37.68 (4) & 35.42 (4) & 46.42 \\
		hu\_szeged & udpipe: self & biaffine ($h_{0}$): self & biaffine ($h_{0}$)*3: self & 82.66 (1) & 80.96 (1) & 79.47 \\
		hy\_armtdp & udpipe: self & biaffine\_trans: self+et\_edt & biaffine\_trans*3: self+et\_edt & 33.90 (3) & 30.87 (3) & 37.01\\
		id\_gsd & uppsala: self & biaffine ($h_{0,1,2}$): self & biaffine ($h_{0,1,2}$)*3: self & 80.05 (1) & 79.19 (1) & 79.13 \\
		it\_isdt & udpipe: self & biaffine ($h_{0}$): self & biaffine ($h_{0,1,2}$)*3: self & 92.00 (1) & 91.71 (1) & 91.47 \\
		it\_postwita & uppsala: self & biaffine ($h_{0,1,2}$): self & biaffine ($h_{0,1,2}$)*3: self+it\_isdt & 79.39 (1) & 78.69 (1) & 78.62 \\
		ja\_gsd & udpipe+scir: self & biaffine ($h_{0}$): self & biaffine ($h_{0}$)*3: self & 83.11 (1) & 82.70 (1) & 79.97 \\
		ja\_modern & udpipe+scir: ja\_gsd & biaffine ($h_{0}$): ja\_gsd & biaffine ($h_{0}$)*3: ja\_gsd & 26.58 (4) &  25.16 (4) & 28.33 \\
		kk\_ktb & udpipe: self & biaffine\_trans: self+tr\_imst & biaffine\_trans*3: self+tr\_imst & 23.92 (10) & 23.18 (13) & 31.93 \\
		kmr\_mg & udpipe: self & biaffine\_trans: self+fa\_seraji & biaffine\_trans*3: self+fa\_seraji & 26.26 (5) & 24.58 (6) & 30.41\\
		ko\_gsd & udpipe: self & biaffine ($h_{0}$): self & biaffine ($h_{0,1,2}$)*3: self & 85.14 (1) & 84.76 (1) & 84.31 \\
		ko\_kaist & udpipe: self & biaffine ($h_{0}$): self & biaffine ($h_{0,1,2}$)*3: self & 86.91 (1) & 86.61 (2) & 86.84 \\
		la\_ittb & udpipe: self & biaffine ($h_{0,1,2}$): self & biaffine ($h_{0,1,2}$)*3: self & 87.08 (1) & 86.50 (2) & 86.54\\
		la\_perseus & udpipe: self & biaffine ($h_{0}$): self+la\_proiel & biaffine ($h_{0,1,2}$)*3: self+la\_proiel & 72.63 (1) & 72.67 (1) & 68.07 \\
		la\_proiel & uppsala: self & biaffine ($h_{0}$): self & biaffine ($h_{0,1,2}$)*3: self & 73.61 (1) & 72.42 (1) & 71.76 \\
		lv\_lvtb & udpipe: self & biaffine ($h_{0}$): self & biaffine ($h_{0}$)*3: self & 83.97 (1) & 83.04 (1) & 81.85 \\
		nl\_alpino & udpipe: self & biaffine ($h_{0}$): self & biaffine ($h_{0,1,2}$)*3: self+nl\_lassysmall & 89.56 (1) & 89.31 (1) & 87.49 \\
		nl\_lassysmall & uppsala: self & biaffine ($h_{0,1,2}$): self & biaffine ($h_{0,1,2}$)*3: self+nl\_alpino & 86.84 (1) & 86.57 (1) & 84.27 \\
		no\_bokmaal & udpipe: self & biaffine ($h_{0,1,2}$): self & biaffine ($h_{0,1,2}$)*3: self & 91.23 (1) & 90.89 (1) & 90.37 \\
		no\_nynorsk & udpipe: self & biaffine ($h_{0}$): self & biaffine ($h_{0,1,2}$)*3: self & 90.99 (1) & 90.62 (1) & 89.46 \\
		no\_nynorsklia & udpipe: self & biaffine ($h_{0}$): self+no\_nynorsk & biaffine ($h_{0,1,2}$)*3: self+no\_nynorsk & 70.34 (1) & 69.06 (1) & 68.71 \\
		pcm\_nsc & udpipe: en\_ewt & biaffine ($h_{0}$): en\_ewt & biaffine ($h_{0,1,2}$)*3: en\_ewt & 24.48 (2) & 25.16 (2) & 30.07 \\
		pl\_lfg & udpipe: self & biaffine ($h_{0}$): self & biaffine ($h_{0,1,2}$)*3: self & 94.86 (1) & 94.63 (1) & 94.62 \\
		pl\_sz & udpipe: self & biaffine ($h_{0}$): self & biaffine ($h_{0,1,2}$)*3: self & 92.23 (1) & 91.67 (1) & 91.59  \\
		pt\_bosque & uppsala: self & biaffine ($h_{0,1,2}$): self & biaffine ($h_{0,1,2}$)*3: self & 87.61 (3) & 87.32 (5) & 87.81 \\
		ro\_rrt & udpipe: self & biaffine ($h_{0}$): self & biaffine ($h_{0,1,2}$)*3: self & 86.87 (1) & 86.07 (3) & 86.33 \\
		ru\_syntagrus & udpipe: self & biaffine ($h_{0,1,2}$): self & biaffine ($h_{0}$)*3: self & 92.48 (1) & 92.26 (1) & 91.72  \\
		ru\_taiga & udpipe: self & biaffine ($h_{0,1,2}$): self+ru\_syntagrus & biaffine ($h_{0,1,2}$)*3: self+ru\_syntagrus & 71.81 (3) & 71.62 (3) & 74.24  \\
		sk\_snk & udpipe: self & biaffine ($h_{0,1,2}$): self & biaffine ($h_{0,1,2}$)*3: self & 88.85 (1) & 88.29 (1) & 87.59 \\
		sl\_ssj & uppsala: self & biaffine ($h_{0}$): self & biaffine ($h_{0,1,2}$)*3: self & 91.47 (1) & 91.08 (2) & 91.26 \\
		sl\_sst & udpipe: self & biaffine ($h_{0}$): self & biaffine ($h_{0,1,2}$)*3: self+sl\_ssj & 61.39 (1) & 59.90 (1) & 58.12 \\
		sme\_giella & udpipe: self & biaffine (none): self & biaffine ($h_{0,1,2}$)*3: self & 69.06 (3) & 67.43 (5) & 69.87  \\
		sr\_set & udpipe: self & biaffine (none): self & biaffine ($h_{0,1,2}$)*3: self & 88.33 (3) & 87.78 (5) & 88.66 \\
		sv\_lines & udpipe: self & biaffine ($h_{0}$): self & biaffine ($h_{0}$)*3: self+sv\_talbanken & 84.08 (1) & 83.64 (1) & 81.97 \\
		sv\_pud & udpipe: sv\_lines & biaffine ($h_{0}$): sv\_lines & biaffine ($h_{0}$)*3: sv\_lines+sv\_talbanken & 80.35 (1) & 79.78 (1) & 79.71  \\
		sv\_talbanken & udpipe: self & biaffine ($h_{0,1,2}$): self & biaffine ($h_{0}$)*3: self+sv\_lines & 88.63 (1) & 88.26 (1) & 86.45 \\
		th\_pud & thai & biaffine\_trans: zh\_gsd & biaffine\_trans*3: zh\_gsd & 0.64 (14) & 0.61 (15) & 13.70 \\
		tr\_imst & udpipe: self & biaffine ($h_{0}$): self & biaffine ($h_{0,1,2}$)*3: self & 66.44 (1) & 64.91 (1) & 64.79 \\
		ug\_udt & udpipe: self & biaffine ($h_{0}$): self & biaffine ($h_{0}$)*3: self & 67.05 (1) & 66.20 (1)& 65.23 \\
		uk\_iu & udpipe: self & biaffine ($h_{0}$): self & biaffine ($h_{0}$)*3: self+ru\_syntagrus & 88.43 (1) & 87.79 (1) & 85.16 \\
		ur\_udtb & udpipe: self & biaffine ($h_{0}$): self & biaffine ($h_{0}$)*3: self & 83.39 (1) & 82.17 (1) & 82.15  \\
		vi\_vtb & udpipe+scir: self & biaffine ($h_{0,1,2}$): self & biaffine ($h_{0}$)*3: self & 55.22 (1) & 53.92 (1) & 47.41  \\
		zh\_gsd & udpipe+scir: self & biaffine (none): self & biaffine ($h_{0,1,2}$)*3: self & 76.77 (1) & 75.55 (1) & 71.04  \\
		\hline
		\textit{average} & & & & 75.84 (1) & 75.26 (1) & 
	\end{tabular}
	\caption{The strategies used in the final submission. 
		The \textit{toolkit} and \textit{model} are separated by colon.
		(\textit{uppsala}: the Uppsala segmentor; \textit{scir}: our segmentor;
		\textit{biaffine}: the biaffine tagger and parser;
		\textit{biaffine\_trans}: our transfer parser for low-resource languages.)
		$h_{0}$ and $h_{0,1,2}$ denotes the ELMo used to train the model.
		$h_0$ means using $\mathbf{h}_{i, 0}^{(LM)}$ and $h_{0,1,2}$ means using $\sum_{j=0}^{2} \mathbf{h}_{i, j}^{(LM)}$.
		\textit{self} denotes that the model is trained with the treebank itself.
		If the model field is not filled with \textit{self}, the model is trained with treebank concatenation.
		The \textit{ref.} column shows the top performing system if we are not top, or the second-best performing system on LAS.
		We also show the results without parser ensemble and our unofficial ranks of this system.
%		A reference system is the top performing system on that metric
	}\label{tbl:big}
\end{table*}

We also study how preprocessing contributes to the final parsing performance.
The experimental results on the development set are shown in Table \ref{tbl:preprocess}.
From this table, the performance of word segmentation
is almost linearly correlated with the final performance.
Similar trends on sentence segmentation performance are witnessed
but \textit{el\_gdt} and \textit{pt\_bosque} presents some exceptions
where better preprocess leads drop in the final parsing performance.

\subsection{Parsing Strategies and Test Set Evaluation}
Using the development set and cross validation,
we choose the best model and data combination and
the choices are shown in Table \ref{tbl:big}
along with the test evaluation.
From this table, we can see that our system gains more improvements
when both ELMo and parser ensemble are used.
For some treebanks, concatenation also contributes to the improvements.
Parsing Japanese, Vietnamese, and Chinese clearly benefits from better word segmentation.
Since most of the participant teams use single parser
for their system, we also remove the parser ensemble and do a post-contest evaluation.
The results are also shown in this table.
Our system without ensemble achieves an macro-averaged LAS
of 75.26, which unofficially ranks the first according to LAS in the shared task.

We report the time and memory consumption.
A full run over the 82 test sets on the TIRA virtual machine \cite{tira}
takes about 40 hours and consumes about 4G RAM memory.

\section{Conclusion}

Our system submitted to the CoNLL 2018 shared task made several improvements
on last year's winning system from \citet{dozat-qi-manning:2017:K17-3},
including incorporating deep contextualized word embeddings,
parser ensemble, and treebank concatenation.
Experimental results on the development set show the effectiveness of our methods.
Using  these techniques, our system achieved an averaged LAS of 75.84\%
and obtained the first place in LAS in the final evaluation.

\section{Credits}

There are a few references we would like to
give proper credit, especially to data providers:
the core Universal Dependencies paper from LREC 2016 \cite{ud},
the UD version 2.2 datasets \cite{ud22testdata}, 
the baseline \textit{udpipe} model released by \citet{udpipe},
the deep contextualized word embeddings code released by \citet{N18-1202},
the biaffine tagger and parser released by \citet{dozat-qi-manning:2017:K17-3},
the joint sentence segmentor and tokenizer released by \citet{delhoneux-EtAl:2017:K17-3},
and the evaluation platform TIRA \cite{tira}.

\section*{Acknowledgments}
We thank the reviewers for their insightful comments, and the HIT-SCIR colleagues for the coordination on the machine usage.
This work was supported by the National Key Basic Research Program of China
via grant 2014CB340503 and the National Natural Science Foundation of China (NSFC)
via grant 61300113 and 61632011.

%\bibliography{tinydb}
%\bibliographystyle{acl_natbib}

\end{document}